\documentclass[a4paper]{article}

\usepackage{INTERSPEECH2022}
\usepackage{amsmath,amssymb,amsfonts}
\usepackage{graphicx}
\usepackage{tabu}
\usepackage{enumitem}
\usepackage{ragged2e}
\usepackage{multirow}

\title{Multiple-hypothesis RNN-T Loss for Unsupervised Fine-tuning and Self-training of Neural Transducer}
\name{Cong-Thanh Do, Mohan Li, and Rama Doddipatla}
\address{
  Cambridge Research Laboratory, Toshiba Europe Limited, Cambridge, U.K.}
\email{\{cong-thanh.do, mohan.li, rama.doddipatla\}@crl.toshiba.co.uk}	
\begin{document}

\maketitle
\begin{abstract}
This paper proposes a new approach to perform unsupervised fine-tuning and self-training using unlabeled speech data for recurrent neural network (RNN)-Transducer (RNN-T) end-to-end (E2E) automatic speech recognition (ASR) systems. Conventional systems perform fine-tuning/self-training using ASR hypothesis as the targets when using unlabeled audio data and are susceptible to the ASR performance of the base model. Here in order to alleviate the influence of ASR errors while using unlabeled data, we propose a multiple-hypothesis RNN-T loss that incorporates multiple ASR 1-best hypotheses into the loss function. For the fine-tuning task, ASR experiments on Librispeech show that the multiple-hypothesis approach achieves a relative reduction of 14.2\% word error rate (WER) when compared to the single-hypothesis approach, on the \small{\texttt{test\_other}} set. For the self-training task, ASR models are trained using supervised data from Wall Street Journal (WSJ), Aurora-4 along with CHiME-4 real noisy data as unlabeled data. The multiple-hypothesis approach yields a relative reduction of 3.3\% WER on the CHiME-4's single-channel real noisy evaluation set when compared with the single-hypothesis approach.  

%We propose a multiple-hypothesis RNN-T loss which is used during unsupervised fine-tuning and self-training to integrate multiple ASR hypotheses into the loss function. The use of multiple ASR hypotheses helps alleviating the impact of ASR errors in the computation of the loss. In unsupervised fine-tuning performed on Librispeech, the multiple-hypothesis unsupervised fine-tuning approach yields 12.7\% and 14.2\% relative word error rate (WER) reductions compared to the conventional unsupervised fine-tuning using single ASR 1-best hypothesis. In self-training experiments performed with Wall Street Journal (WSJ), Aurora-4, and CHiME-4 data, the multiple-hypothesis self-training yielded 6.0\% and 3.3\% relative WER reductions compared to the single-hypothesis self-training on Aurora-4's and CHiME-4's test data, respectively.
\end{abstract}

\vspace{-5pt}
\section{Introduction}
\label{sec:introduction}

RNN-T \cite{graves2012}, or neural transducer \cite{battenberg2017},  is a popular E2E ASR model which was found successful in both research and industry environments \cite{li_asru_2019, li_interspeech_2020_1}. This model provides state-of-the-art performance in a wide variety of streaming applications \cite{li_interspeech_2020_1}. RNN-T extends connectionist temporal classification (CTC) to define a distribution over output sequences of all lengths, and jointly models both input-output and output-output dependencies \cite{graves2012}. The modeling in RNN-T overcomes the issue regarding the conditional independence between output predictions at different time steps given aligned inputs, introduced in CTC \cite{graves2006}. 

While RNN-T model is widely applied, it would be useful to be able to efficiently customize this model to new application domains \cite{li_interspeech_2020_2}. However, speech data with manual transcriptions are not always available because the cost to obtain the labels is substantially high. How to leverage large-scale unlabeled speech data to improve ASR performance has been a longstanding research problem. Pseudo-labeling \cite{lee_icmlworkshop_2013} is one of the approaches that can be used to customize RNN-T models to new domains using unlabeled data. The pseudo-labeling process starts with training a base model using available labeled data. The base model is then used to label the unlabeled data. The pseudo-labeled data are subsequently used to fine-tune or re-train the base model. The pseudo-labeling process can be repeated to improve the quality of the customized model. Pseudo-labeling has been a practically useful and extensively studied technique in ASR \cite{kahn_icassp_2020, synnaeve_icmlworkshop_2020}. 

In this paper, we present a new approach for unsupervised fine-tuning and self-training of RNN-T model using unlabeled speech data and multiple ASR 1-best hypotheses. Multiple ASR 1-best hypotheses are used instead of a single ASR 1-best hypothesis which is conventionally used in pseudo-labeling. The ASR 1-best hypotheses are incorporated in the RNN-T loss function in order to alleviate the influence of errors in the ASR hypotheses to the computation of the RNN-T loss. This proposal is consistent with our previous work where multiple ASR 1-best hypotheses were incorporated into the CTC loss function to improve the adaptation of CTC-based E2E ASR \cite{do_icassp_2021}. In this respect, a multiple-hypothesis RNN-T loss is proposed to incorporate multiple ASR 1-best hypotheses during unsupervised fine-tuning and self-training. Experiments are performed using data from Aurora-4 \cite{parihar2002}, CHiME-4 \cite{vincent2017}, Wall Street Journal (WSJ) \cite{paul1992}, and Librispeech \cite{panayotov_icassp_2015} corpora. In both unsupervised fine-tuning and self-training scenarios, the approach using multiple ASR 1-best hypotheses is effective in reducing the WER and it outperforms the conventional approach using single ASR 1-best hypothesis.

The paper is organized as follows. Section \ref{sec:related_work} presents related works. The multiple-hypothesis RNN-T loss used for fine-tuning and self-training is introduced in section \ref{sec:mh_rnnt_loss}. Details of the unsupervised fine-tuning and self-training experiments together with the experimental results are presented in sections \ref{sec:unsupervised_adaptation} and \ref{sec:semi-supervised_adaptation}, respectively. Finally, section \ref{sec:conclusion} concludes the paper.

%RNN-T also marginalizes over all possible alignments, like CTC does, while extending CTC by additionally modelling the dependencies between outputs at different time steps. 

\vspace{-5pt}
\section{Related works}
\label{sec:related_work}
\vspace{-5pt}

Research on neural transducer and the customization of this model is an active research topic in ASR \cite{li_interspeech_2020_2, bell2021}. In \cite{huang_interspeech_2020}, the authors investigated which components of the RNN-T model is the most effective one to be adapted when the amount of adaptation data is limited, in supervised and unsupervised speaker adaptation. In \cite{saon_icassp_2021}, i-vector speaker embedding was applied for speaker adaptation in RNN-T E2E ASR. In \cite{pylkkonen_interspeech_2021}, a small amount of textual data was used to adapt the RNN-T's prediction network to obtain fast adaptation of RNN-T model.

Multiple hypotheses were previously used in cross-system acoustic model adaptation where the transcriptions for adaptation were generated by several systems, which were built with various phoneme sets or acoustic front-ends \cite{giuliani2007, stueker2006, gibson2006}. Recently, hypotheses from the $N$-best list are increasingly used in ASR applications. For instance, in \cite{guo_interspeech_2020}, the hypotheses in $N$-best list were used during the minimum WER training of RNN-T E2E ASR. In \cite{moritz_icassp_2021}, an $N$-best list of pseudo-labels were used to generate lattice supervision in self-training using graph-based temporal classification objective function. 

In the present paper, we propose to perform unsupervised fine-tuning and self-training of RNN-T E2E model using unlabeled data and multiple ASR 1-best hypotheses from different ASR systems. In fact, it was proved that better results are obtained when the adaptation process exploits a hypothesis generated by a system different than the one under adaptation \cite{giuliani2007}, similar to in systems combination \cite{meinedo_icslp_2000}. In our approach, the ASR 1-best hypotheses are incorporated into the RNN-T loss functions during fine-tuning and self-training.

%The hybrid autoregressive transducer (HAT) model \cite{variani_icassp_2020} provides a way to measure the quality of the internal language model that can be used to decide whether inference with an external language model is beneficial or not.

\vspace{-5pt}
\section{Multiple-hypothesis RNN-T loss}
\label{sec:mh_rnnt_loss}

Given a $T$-length acoustic feature vector sequence $X = \{\mathbf{x}_t \in \mathbb{R}^d | t = 1,...,T\}$, where $\mathbf{x}_t$ is a $d$-dimensional feature vector at frame $t$, and a transcription $C = \{c_l \in \mathcal{U}| l = 1,...,L\}$ which consists of $L$ characters, where $\mathcal{U}$ is a set of distinct characters, during the training of the RNN-T E2E ASR, the RNN-T loss function $L_{\texttt{RNN-T}}$ is minimized. $L_{\texttt{RNN-T}}$ is defined as follows:

\vspace{-5pt}
\begin{equation}
\label{equ:rnnt_loss_1}
L_{\texttt{RNN-T}} = -\log P_{\theta} (C|X),
\end{equation}

\noindent where $\theta$ are the network parameters and $P_{\theta} (C|X) = \sum_{\widehat{C}}P(\widehat{C}|X), \widehat{C} \in \mathcal{B}^{-1}(C)$. In equation (\ref{equ:rnnt_loss_1}), $C$ is the transcription of $X$, which can either be a manual transcription or an ASR hypothesis, and $\mathcal{B}^{-1}(C)$ is the set of all possible alignments, including the blank symbol, between the transcription $C$ and the input feature vector sequence $X$. In RNN-T, the posterior probability $P(\widehat{C}|X)$ is computed as:

\begin{equation}
\label{equ:rnnt_loss_2} 
P(\widehat{C}|X) = \prod_{\hat{c}_i \in \widehat{C}} P(\hat{c}_i|\mathbf{x}_1...\mathbf{x}_{t_{i}}, \hat{c}_1...\hat{c}_{i-1}). 
\end{equation}

\noindent From equations (\ref{equ:rnnt_loss_1}) and (\ref{equ:rnnt_loss_2}), we have:

\vspace{-10pt}
\begin{equation}
\small
\label{equ:rnnt_loss_3}
L_{\texttt{RNN-T}} = - \log \left ( \sum_{\widehat{C} \in \mathcal{B}^{-1}(C)} \prod_{\hat{c}_i \in \widehat{C}} P(\hat{c}_i|\mathbf{x}_1...\mathbf{x}_{t_{i}}, \hat{c}_1...\hat{c}_{i-1}) \right ).
\end{equation}

It can be observed in equations (\ref{equ:rnnt_loss_2}) and (\ref{equ:rnnt_loss_3}) that both input-output and output-output dependencies are modeled in the RNN-T loss, as mentioned in section \ref{sec:introduction}. In this paper, we propose multiple-hypothesis RNN-T loss $L^{*}_{\texttt{RNN-T}}$ to be used during fine-tuning and self-training as follows:

\begin{equation}
\label{equa:multiple_hypothesis_rnnt}
L^{*}_{\texttt{RNN-T}} = -\left( \sum^{M}_{i=1} \log P_{\theta} (\bar{C}_i|X) \right),
\end{equation}

\noindent where $\bar{C}_i, i = 1, ..., M$ are the ASR 1-best hypotheses obtained by decoding the unlabeled speech data, which is available for fine-tuning or self-training, using $M$ different base models. By combining multiple 1-best hypotheses in the computation of the RNN-T loss, the influence of ASR errors in the 1-best hypotheses to the computation of the RNN-T loss function could be alleviated \cite{do_icassp_2021}.

In this work, the unlabeled speech data available for unsupervised fine-tuning and self-training are assumed to have no manual transcriptions. The proposed multiple-hypothesis RNN-T loss is used in both unsupervised fine-tuning and self-training to incorporate multiple ASR 1-best hypotheses.

%To make beam search effective, the conditional independence assumption is artificially broken by the inclusion of an external language model \cite{battenberg2017}, and decoding is then the task of finding the argmax of 
%\begin{equation}
%\log (P_{CTC}(y|x)) + \alpha \log (P_{LM}(y)) + \beta wordcount(y)
%\end{equation}

%This decoding is approximate, and performed using beam search, typically with a large beam or lattice. The above equation presents a discrepancy between how these models are trained and tested. To address this, model could be further fine-tuned with a loss function that also incorporates language model like sMBR, but the main issue is still the absence of dependence between predictions.

%\section{Adaptation experiments}
%\label{sec:adaptation_experiments}

%In this work, unsupervised and semi-supervised adaptations are studied. The adaptation speech data are assumed to have no manual transcriptions. The proposed multiple-hypothesis %RNN-T loss is used in both unsupervised and semi-supervised adaptations to incorporate multiple ASR hypotheses during adaptation.

\vspace{-5pt}
\section{Unsupervised fine-tuning}
\label{sec:unsupervised_adaptation}

\subsection{Method}
\label{sec:unsupervised_adaptation_theory}

In unsupervised fine-tuning experiments, we assume that labeled speech data is available for training a RNN-T base model. The base model is normally evaluated on test data which includes audio only. During unsupervised fine-tuning, the audio of the test data is used to fine-tune the base model. Since the labels of the test audio are not available, the base model is used to decode the test audio. The 1-best hypotheses obtained from the decoding are used as the labels for the test audio data during the fine-tuning of the base model.

During the unsupervised fine-tuning, back-propagation algorithm is used to fine-tune all parameters of the base model with respect to the loss function, using the available data. Stochastic gradient descent (SGD) algorithm \cite{ruder2016} is used during the minimization of the loss function. In conventional unsupervised fine-tuning, the standard RNN-T loss $L_{\texttt{RNN-T}}$ (see equation (\ref{equ:rnnt_loss_1})) and one single ASR 1-best hypothesis are used while in the multiple-hypothesis unsupervised fine-tuning, the proposed multiple-hypothesis RNN-T loss $L^{*}_{\texttt{RNN-T}}$ (see equation (\ref{equa:multiple_hypothesis_rnnt})) and multiple ASR 1-best hypotheses are used. During the training of the base model, manual transcriptions of audio data are used in the standard RNN-T loss $L_{\texttt{RNN-T}}$.

\vspace{-5pt}
\begin{figure}[ht]
	\centering
		\includegraphics[width=0.95\columnwidth]{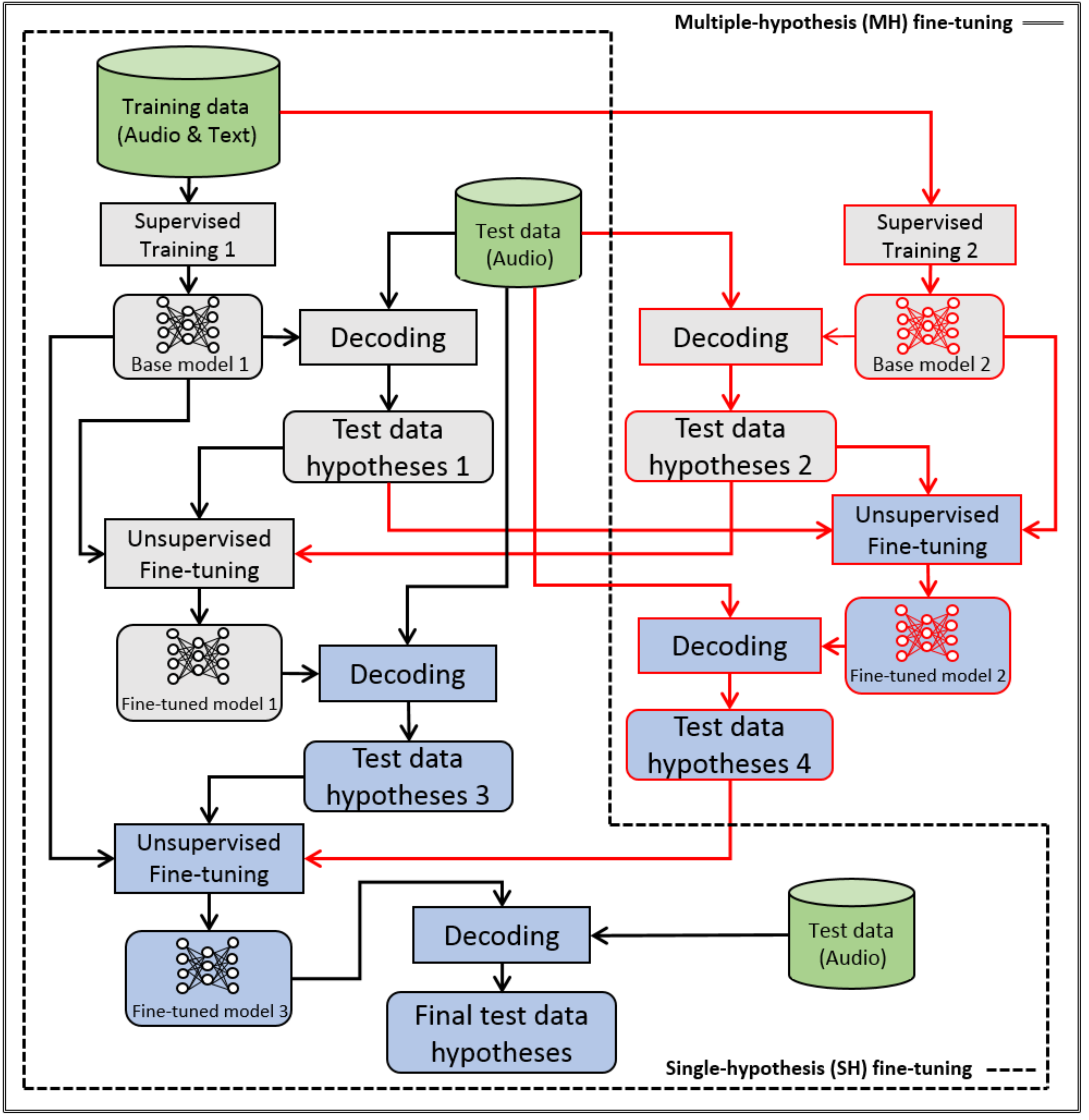}
	\caption{\label{fig:unsupervised_adaptation} Unsupervised fine-tuning of RNN-T model using unlabeled audio data and multiple ASR 1-best hypotheses. The blocks surrounded by the dashed lines are used in conventional single-hypothesis fine-tuning. The red outline blocks and arrows are used only for multiple-hypothesis fine-tuning. The blocks filled with gray color are used in the first iteration of fine-tuning while those filled with pastel blue color are used in the second fine-tuning iteration. Arrows connecting the test audio data and unsupervised fine-tuning blocks are not shown to reduce the density of connections and improve clarity.\vspace{-10pt}}
\end{figure}

The unsupervised fine-tuning experiments are depicted in Fig. \ref{fig:unsupervised_adaptation}. The processing used by the conventional single-hypothesis unsupervised fine-tuning are surrounded by the dashed lines. Additional processing are required by the unsupervised fine-tuning using multiple ASR hypotheses and these processing are represented in red color. In this work, from the labeled training data we train different RNN-T base models, in addition to the one that needs to be fine-tuned, to decode the test audio data to obtain multiple ASR 1-best hypotheses. 

In Fig. \ref{fig:unsupervised_adaptation}, the ``Unsupervised Fine-tuning'' blocks can receive either single or multiple ASR hypotheses. When single hypothesis is received, the standard RNN-T loss is used. When multiple hypotheses are received, the multiple-hypothesis RNN-T loss is used to incorporate theses hypotheses. In the fine-tuning using single hypothesis, the additional hypotheses, represented by red arrows in Fig. \ref{fig:unsupervised_adaptation}, are not used. Both single-hypothesis and multiple-hypothesis fine-tuning can do more than one iteration to improve further the performance. Indeed, the fine-tuned model can be used to decode the test audio data again to obtain better hypotheses. This process can be repeated until the performance improvement is saturated. In this work, depending on the experiment, we perform either one or two fine-tuning iterations, as shown in Fig. \ref{fig:unsupervised_adaptation}.  

\subsection{Data and model architecture}
\label{sec:unsupervised_adaptation_experiments}

The unsupervised fine-tuning experiments are carried out with Librispeech corpus \cite{panayotov_icassp_2015}. The Librispeech dataset contains approximately 1000 hours of speech data that is derived from read English audiobooks. The training data is split into 3 subsets, including two ``clean'' subsets with the duration of 100 and 360 hours, as well as an ``other'' subset with the duration of 500 hours. While the dev and test sets are also split into the ``clean'' and ``other'' categories with the duration of 5 hours and 20 minutes each, facilitating evaluation under different speech conditions. The total number of speakers in the training and dev/test sets are 2338 and 77, respectively. In our experiment, we used the full training data (960 hours) to train the ASR models. The dev set used for the training is a combination of the ``clean'' and ``other'' dev sets.

A word-level language model (LM) is trained to rescore the hypotheses produced by the RNN-T ASR system. The architecture of the LM is a 4-layer long short-term memory (LSTM) network \cite{hochreiter1997}, with each layer having 2048 memory blocks. We use the transcriptions of the Librispeech’s training set, as well as the texts of 14500 public domain books as the training data, and the vocabulary size is 89112. The LM is trained with SGD optimizer for 4 data epochs until it fully converges. This LM is used in all the decoding in the present paper.

The architecture of the RNN-T base model is as follows. The encoder consists of 6 bidirectional LSTM layers with 1024 memory blocks per layer per direction. The joint network is a feed-forward linear network which projects the 1024-dimensional stacked encoder vectors from the last encoder layer. The decoder is a 1-layer LSTM which contains 1024 memory blocks. Dropout \cite{srivastava2014} is applied in the decoder network. The RNN-T's output units are English characters. The use of characters as output units helps to simplify and generalize the training of the RNN-T base models even though these models might not be as strong as those using context-dependent phonetic units, for instance word-pieces, as output units \cite{irie_interspeech_2019}. %While using characters as output units of the RNN-T models might not yield a performance as strong as that of the models using context-dependent phonetic units, for instance word-pieces, as output units \cite{irie_interspeech_2019}, the training of the RNN-T models is simplified and is straightforward to generalize.%

40-dimensional filter-bank are extracted as acoustic features. The filter-bank features are appended with 3-dimensional pitch features. Delta and acceleration features are computed from static features. Static and dynamic features are appended in one feature vector of 129 dimensions. The feature extraction is performed by using Kaldi toolkit \cite{povey2011}. The model training is done with 20 epochs and a beam size of 20 is used in the decoding. Training and decoding are performed by using ESPnet toolkit \cite{watanabe2018} and Pytorch \cite{pytorch_neurips2019}.

\vspace{-5pt}
\subsection{Results}
\label{sec:unsupervised_adaptation_results}

\subsubsection{Multiple hypotheses with comparable WERs}
\label{sec:unsupervised_adaptation_results_1}

We examine the effectiveness of the multiple-hypothesis unsupervised fine-tuning when multiple hypotheses with comparable WERs are used. These hypotheses can be obtained by creating a variant of the base model and use the variant model to decode the test audio data to obtain additional hypotheses. In this work, we simply change the dropout rate in the RNN-T's decoder network to train the ``Base model 2'' which is a variant of the ``Base model 1'' (see Fig. \ref{fig:unsupervised_adaptation}). Decoding the test data with the base model 2 produces a variant of the ASR hypotheses compared to that produced by the base model 1 \cite{dey_interspeech_2019, khurana_icassp_2021}. Their WERs can be seen in the first two lines of Table \ref{tab:rnnt_unsupervised_adaptation_librispeech}. 

Table \ref{tab:rnnt_unsupervised_adaptation_librispeech} shows experimental results when two hypotheses are used in the multiple-hypothesis unsupervised fine-tuning. The results are obtained with the fine-tuned model 3 (see Fig. \ref{fig:unsupervised_adaptation}), after two fine-tuning iterations. The base model 2 applies a dropout rate of 0.5 while the base model 1 applies a dropout rate of 0.1 in the decoder network. It can be seen in Table \ref{tab:rnnt_unsupervised_adaptation_librispeech} that both the unsupervised fine-tunings using single hypothesis (SH) and multiple hypotheses (MH) yield WER reductions compared to the base models. The MH unsupervised fine-tuning outperforms the conventional SH unsupervised fine-tuning for both \small{\texttt{test\_clean}} and \small{\texttt{test\_other}} sets. The relative WER reductions obtained by the MH fine-tuning compared to the SH fine-tuning on \small{\texttt{test\_clean}} and \small{\texttt{test\_other}} sets are 12.7\% and 14.2\%, respectively.

\vspace{-5pt}
\begin{table}[ht]
\centering
\caption{Unsupervised fine-tuning experiments on Librispeech. SH denotes single hypothesis and MH denotes multiple hypotheses. Two ASR hypotheses are used in the fine-tuning using multiple ASR hypotheses.\vspace{-5pt}}
\label{tab:rnnt_unsupervised_adaptation_librispeech}
\begin{tabu}{|l|c|c|}
\hline
\multirow{2}{*}{\hspace{20pt} Fine-tuning method}  & \multicolumn{2}{c|}{WER (in \%)} \\ \cline{2-3} 
&  												Test\_clean    & Test\_other           \\ \hline \hline
Base model 1                           & 8.5             & 22.9  \\ \hline
Base model 2                           & 8.1             & 23.1  \\ \hline
SH fine-tuning (1 hypothesis)       & 7.9             & 22.6  \\ \hline
MH fine-tuning (2 hypotheses)     & 6.9             & 19.4  \\ \hline
\end{tabu}
\end{table}

%\begin{table}[ht]
%\centering
%\caption{Unsupervised adaptation experiments on Librispeech.}
%\label{tab:rnnt_unsupervised_adaptation_librispeech}
%\begin{tabu}{|l|c|c|}
%\hline
%\multirow{2}{*}{\hspace{10pt} Adaptation method}  & \multicolumn{2}{c|}{WER (in \%)} \\ \cline{2-3} 
%&  												Test\_clean    & Test\_other           \\ \hline \hline
%Initial model (no adaptation)                           & 8.5             & 22.9  \\ \hline
%SH adaptation (1st pass)      & 8.2             & 22.8  \\ \hline
%SH adaptation (2nd pass)       & 7.9             & 22.6  \\ \hline
%MH adaptation (1st pass)     & 7.2             & 20.0  \\ \hline
%MH adaptation (2nd pass)     & 6.9             & 19.4  \\ \hline
%\end{tabu}
%\end{table}

\vspace{-10pt}
\subsubsection{Multiple hypotheses with different WERs}

We also examine the effectiveness of the multiple-hypothesis fine-tuning when the additional model used to decode the test audio data has a substantial lower WER than the model that needs to be fine-tuned. This situation can happen when, for instance, a model from one site is used to help fine-tuning a model at another site and the performance of the two or more models are quite different.

In this experiment, a second hypothesis with substantially lower WER compared to the one produced by the model being fine-tuned is used. By using this setting, we verify whether the fine-tuned model can achieve lower WER compared to the base models participated in the fine-tuning. In this experiment, we perform one iteration of fine-tuning instead of two.% Experiments to verify whether the MH adaptation can benefit from ASR hypotheses with higher WER compared to that of the to-be-adapted system will be presented in section \ref{sec:semi-supervised_adaptation}. 

\begin{table}[ht]
\centering
\caption{Unsupervised fine-tuning experiments on Librispeech when 2 and 4 hypotheses are used in the fine-tuning using multiple ASR 1-best hypotheses.\vspace{-5pt}}
\label{tab:rnnt_unsupervised_adaptation_librispeech_multipleHyps}
\begin{tabu}{|l|c|c|}
\hline
\multirow{2}{*}{\hspace{20pt} Fine-tuning method}  & \multicolumn{2}{c|}{WER (in \%)} \\ \cline{2-3} 
&  												Test\_clean    & Test\_other           \\ \hline \hline
Base model 1                           & 8.5             & 22.9  \\ \hline
Base model 3                           & 6.8             & 20.0  \\ \hline \hline
MH fine-tuning (2 hypotheses)     & 6.6             & 18.5  \\ \hline \hline
Base model 2                           & 8.1             & 23.1  \\ \hline
Base model 4                           & 7.0             & 20.7  \\ \hline \hline
MH fine-tuning (4 hypotheses)     & 5.7             & 17.2  \\ \hline
\end{tabu}
\end{table}

Experimental results are shown in Table \ref{tab:rnnt_unsupervised_adaptation_librispeech_multipleHyps}. In the fine-tuning using 2 hypotheses, we replace the hypotheses produced by the base model 2 used in Table \ref{tab:rnnt_unsupervised_adaptation_librispeech} by the hypotheses produced by another model, the base model 3. The base model 1 is the model to be fine-tuned. The hypotheses produced by the base model 3 have substantial lower WER compared to those produced by the base model 1. The base model 3 is trained similarly to the base model 1 where the dropout rates in the decoder networks of both models equal 0.1, but with 3-fold speed perturbation \cite{ko_interspeech_2015} applied during training. The base model 1 is trained on 960 hours of Librispeech training data without speed perturbation. It can be seen in Table \ref{tab:rnnt_unsupervised_adaptation_librispeech_multipleHyps} that the MH fine-tuning using two hypotheses creates a fine-tuned model which yields lower WERs on \small{\texttt{test\_clean}} and \small{\texttt{test\_other}} sets compared to the base model 3.

We include two additional hypotheses, one produced by the base model 2, and another produced by a base model 4 which is trained similarly to the base model 3 but with the decoder network's dropout rate equals 0.5. When including all 4 hypotheses in the fine-tuning using multiple hypotheses, we obtain further WER reductions on both \small{\texttt{test\_clean}} and \small{\texttt{test\_other}} sets. The fine-tuned model, obtained by the fine-tuning using 4 hypotheses, yields 16.2\% and 14.0\% relative WER reductions on \small{\texttt{test\_clean}} and \small{\texttt{test\_other}} sets, respectively, compared to the best base model (base model 3).  

\vspace{-5pt}
\section{Self-training}
\label{sec:semi-supervised_adaptation}
\vspace{-5pt}

\subsection{Method}
\label{sec:semi-supervised_adaptation_theory}

\begin{figure}[ht]
	\centering
		\includegraphics[width=0.95\columnwidth]{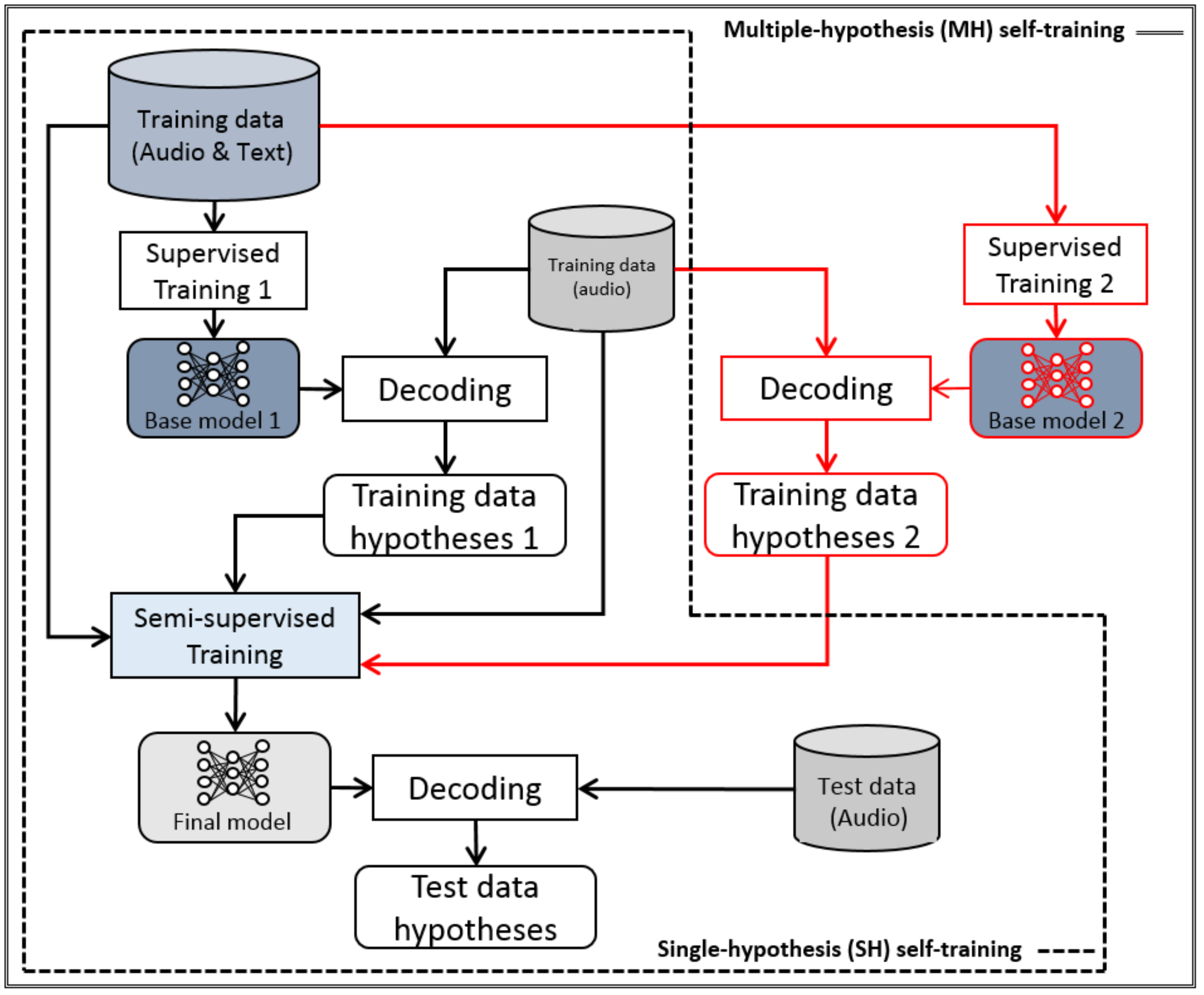}
	\caption{\label{fig:semi-supervised_adaptation} Self-training of RNN-T model using additional unlabeled audio data and multiple ASR 1-best hypotheses. The blocks surrounded by the dashed lines are used in conventional single-hypothesis self-training. The red outline blocks and arrows are used only for multiple-hypothesis self-training.\vspace{-10pt}}
\end{figure}

Similar to the unsupervised fine-tuning experiment, in self-training, training data with manual transcriptions are available to train a base model. We study the scenario where additional audio data for training is available but the manual transcriptions, or labels, of these data are not available. In conventional self-training using pseudo-labeling, a base model (base model 1, see Fig. \ref{fig:semi-supervised_adaptation}) is trained using the supervised audio training data. This model is then used to decode the additional training data. In this work, the base model 2 is trained similarly to the base model 1. The only difference between these models is the dropout rate of the decoder network which equals 0.1 in the base model 1 and 0.5 in the base model 2. The additional audio data and ASR 1-best hypotheses are then combined with the supervised training data to do semi-supervised training of the final model.

In self-training using multiple ASR hypotheses, a variant of the base model is trained using the supervised training data to create the base model 2 (see Fig. \ref{fig:semi-supervised_adaptation}). This model is used to decode the additional training data and the obtained hypotheses are used in the multiple-hypothesis semi-supervised training. The SH and MH self-training are depicted in Fig. \ref{fig:semi-supervised_adaptation}.

\vspace{-5pt}
\subsection{Data and system}
\label{sec:semi-supervised_adaptation_experiments}

We use the WSJ's clean training data and Aurora-4's multi-condition training data to train the base models, with 96 hours of speech in total. Speed perturbation is applied to increase the training data to 288 hours. Around 9 hours of dev data is taken from Aurora-4's dev data and is used during training. The system is evaluated on two test sets, one of 4.8 hours extracted from Aurora-4's 14 test sets, and one of 2.2 hours which is the CHiME-4's single-channel real noisy evaluation set, \small{\texttt{et05\_real\_isolated\_1ch\_track}}. 17.5-hour additional training data is taken from the \small{\texttt{tr05\_real\_noisy}} training set of CHiME-4 which consists of 9600 utterances recorded in real noisy environments \cite{vincent2017}. When being added to the original training data, speed perturbation is applied to the additional training data to increase the amount of the added training data to 52.5 hours. The RNN-T model in this experiment has the same architecture and specifications as that used in section \ref{sec:unsupervised_adaptation_experiments}. The RNN-LM mentioned in section \ref{sec:unsupervised_adaptation_experiments} is used in the decoding. The WERs of the base model 1 and base model 2 on the additional training data are 52.9\% and 54.7\%, respectively.

\vspace{-5pt}
\subsection{Results}
\label{sec:semi-supervised_adaptation_results}
\vspace{-5pt}

Table \ref{tab:rnnt_semisupervised_adaptation_aurora4} shows the results of the self-training experiments. The results show that both SH and MH self-training yield WER reductions on the test data from Aurora-4 and CHiME-4. The MH self-training yields larger WER reductions compared to the SH self-training. On Aurora-4 test data, the MH self-training yields 6.8\% and 6.0\% relative WER reductions compared to the base model 1, which is the better base model, and the SH self-training, respectively. On CHiME-4 evaluation data, the MH self-training yields 7.8\% and 3.3\% relative WER reductions compared to the base model 1 and the SH self-training, respectively. The ``Supervised training'' in which the manual transcriptions of the additional training data are used yields the lowest WERs on both Aurora-4 and CHiME-4 test data, as can be seen in Table \ref{tab:rnnt_semisupervised_adaptation_aurora4}.

\vspace{-5pt}
\begin{table}[ht]
\centering
\caption{Self-training experiments using training data from WSJ, Aurora-4, and CHiME-4. Evaluation data are from Aurora-4 and CHiME-4.\vspace{-5pt}}
\label{tab:rnnt_semisupervised_adaptation_aurora4}
\begin{tabu}{|l|c|c|}
\hline
\multirow{2}{*}{\hspace{1pt} Self-training method}  & \multicolumn{2}{c|}{WER (in \%)} \\ \cline{2-3} 
&  												Aurora-4    & CHiME-4           \\ \hline \hline
Base model 1                           & 23.6             & 60.4  \\ \hline
Base model 2                           & 24.1             & 61.1  \\ \hline
SH self-training       & 23.4             & 57.6  \\ \hline
MH self-training     & 22.0             & 55.7  \\ \hline
Supervised training             & 21.0              & 44.2  \\ \hline
\end{tabu}
\end{table}

\vspace{-10pt}
\section{Conclusion}
\label{sec:conclusion}
\vspace{-5pt}

We have proposed a new approach to perform unsupervised fine-tuning and self-training of RNN-T E2E ASR model. The proposed approach used unlabeled speech data and multiple ASR 1-best hypotheses. A multiple-hypothesis RNN-T loss was proposed to incorporate multiple ASR 1-best hypotheses into the loss function which was used during fine-tuning and self-training. In the fine-tuning experiments performed with Librispeech, the approach using multiple ASR 1-best hypotheses yielded 14.2\% relative WER reductions compared to the conventional single-hypothesis approach, on the \small{\texttt{test\_other}} set. In the self-training experiments performed with WSJ, Aurora-4, and CHiME-4 data, the multiple-hypothesis approach yielded 3.3\% relative WER reduction compared to the conventional single-hypothesis approach, on the CHiME-4's single-channel real noisy evaluation set. 

These results show that incorporating multiple ASR 1-best from different systems in the RNN-T loss is beneficial in alleviating the influence of ASR errors and helps improving the unsupervised fine-tuning and self-training performance. Future works will explore new ways of integrating multiple ASR hypotheses into the loss and solutions to reduce the computational complexity of the current approach, including the use of $N$-best hypotheses from one ASR system.

\bibliographystyle{IEEEtran}

\bibliography{mybib}

\end{document}